\documentclass{article}
\PassOptionsToPackage{numbers,compress}{natbib}
\usepackage[preprint]{neurips_2025}
\usepackage[utf8]{inputenc}
\usepackage[T1]{fontenc}
\usepackage{microtype}
\usepackage{amsmath,amssymb}
\usepackage{booktabs}
\usepackage{tabularx}
\usepackage{enumitem}
\usepackage{xcolor}
\usepackage{colortbl}
\usepackage{graphicx}
\usepackage{float}

\definecolor{nAccent}{HTML}{BB480B}
\definecolor{nInk}{HTML}{222222}
\definecolor{nMuted}{HTML}{6B5B4A}
\definecolor{nBorder}{HTML}{D6CFC0}
\definecolor{nBorderLight}{HTML}{EDE6DA}

\usepackage[font=small,labelfont={bf,color=nAccent},labelsep=period]{caption}
\usepackage[colorlinks=true,linkcolor=nAccent,citecolor=nAccent,urlcolor=nAccent]{hyperref}
\usepackage{placeins}
\arrayrulecolor{nBorder}

\newcommand{\thead}[1]{\textbf{#1}}
\newcommand{\rowrule}{\arrayrulecolor{nBorderLight}\specialrule{0.4pt}{3pt}{3pt}\arrayrulecolor{nBorder}}

\setlist{itemsep=2pt,topsep=3pt}
\newcommand{\code}[1]{\texttt{\small #1}}

\makeatletter\let\rawinput\@@input\makeatother 

\title{Lean Census: A Distributional Analysis of AI-Generated Lean Code\thanks{We used AI
  assistants, including Claude, to help write the analysis code and edit the text.
  The authors checked all results and take responsibility for the paper.}}
\author{%
  Justin Asher$^{1}$\hspace{1.4em}Pietro Pellegrino$^{1,2}$ \\[1.2ex]
  \small\normalfont $^{1}$Project Numina \\[0.3ex]
  \small\normalfont $^{2}$University of Naples Federico II \\[0.8ex]
  \small\normalfont \texttt{justin@projectnumina.ai}\hspace{1.4em}\texttt{pietro@projectnumina.ai}
}

\begin{document}
\maketitle

\begin{abstract}
We compare two human-written and four AI-generated Lean libraries with Mathlib
through the distributions of simple statistics of their code. For each library
we measure inline claims in proofs, proof length, how often each declaration is
used, and design choices such as instances and private declarations, and we read
each against the variation among Mathlib's own areas. The
AI-generated libraries have more inline claims per theorem, longer proofs,
fewer instances, and more private declarations than the human-written ones. We
release the code for the analysis as a package, \code{lean-census}, at
\url{https://github.com/project-numina/lean-census}.
\end{abstract}

\section{Introduction}

AI systems now write Lean libraries with millions of lines of code and little
human review. A library is useful only if others can build on it. We compare the distributions of simple statistics over
the code of human-written and AI-generated Lean libraries. We call a library human-written when people review
every contribution, as in Mathlib~\cite{mathlib2019}, even if some of it was
drafted with AI tools. We call it AI-generated when AI systems write it with little or no line-by-line
human review.

We study six libraries against Mathlib (Table~\ref{tab:corpora}). Two are
human-written, CSLib~\cite{barrett2026cslib} and
PhysLib~\cite{toobysmith2025heplean,physlib2026library}, and four are
AI-generated, FormalFrontier~\cite{formalfrontier2026libraries}, Tau
Ceti~\cite{tauceti2026library}, CSLib Frontier~\cite{cslibfrontier2026library},
and Atlas~\cite{rammal2026atlasscale,atlas2026repository}. Tau Ceti alone has
nearly three times as many lines as Mathlib. Other large AI-generated projects
include Anthropic's Fermat's Last Theorem project~\cite{anthropic2026flt} and
Axiom's formalizations of the classification of finite simple
groups~\cite{axiom2026cfsg} and of the Stacks Project~\cite{axiom2026stacks}.
We leave these out because they are expensive to build and analyze, and we
assume smaller libraries suffice for our questions.

Audits of AI-generated Lean have so far relied mostly on judgment. An expert
reviews a project's definitions, file organization, and
interfaces~\cite{ilin2026sorries}, or a language model flags unnecessary
hypotheses and duplicated material one declaration or file at a
time~\cite{morrison2026simple,degenne2026indicators}. A few static counts, such
as how often definitions are unfolded, cover whole
libraries~\cite{degenne2026indicators}.
These audits find concrete problems, but none reads the distributions of
statements, proofs, and dependencies across a library against the variation
within human-written code.

Our premise is that if AI-generated code is structured like human-written code,
the two should have similar distributions. Proving
intermediate facts as small lemmas, rather than with \code{have} inside a long
proof, makes them reusable. A library built this way has short statements and proofs,
and its declarations are used many times. This is compression in the sense of \citet{aksenov2026compression}, where each fact is
written once and referred to many times.

We find that every AI-generated library lies outside the range of the three
human-written libraries on inline claims, proof length, instances,
private declarations, and explicit arguments. AI-generated proofs state more intermediate facts, with 0.81 to 3.22 inline claims per theorem
against 0.21 to 0.64. Their median proofs are 22 to 49 tokens long against 10 to
17, and 0.8 to 3.2\% of their declarations are instances against 5.4 to 10.7\%.
They mark 14 to 37\% of their declarations private, meaning that no other file
can use them, against 1 to 4\%. Part of the difference in proof structure may reflect purpose rather than
authorship, since Mathlib's own archive of one-off proofs is structured much like
the AI-generated libraries. On instances,
private declarations, and explicit arguments, however, the archive is like
Mathlib (Section~\ref{sec:purpose}).

\section{Mathematics as code}
\label{sec:background}

Formal mathematics is software, since by the Curry--Howard
correspondence a proof is a
program~\cite{howard1980formulae,wadler2015propositions}. As
\citet{martinlof1984constructive} wrote, ``it no longer seems possible to
distinguish the discipline of programming from constructive mathematics.'' We therefore draw on software engineering, which has long studied how to divide
programs into parts, to ask how mathematical libraries should be
structured~\cite{gonthier2013software}.
\citet{avigad2020modularity} uses the same principles to explain how informal
mathematics is organized, and \citet{commelin2024abstraction} argue that
mathematicians already rely on abstraction boundaries, which formalization
makes explicit.

\paragraph{Modules and interfaces.} \citet{parnas1972criteria} argued that a
program should be divided into modules that each hide one design decision
behind a small interface. In Lean, a definition and the basic results about it
form such a module~\cite{carette2014realms}. As an example, Mathlib builds the
real numbers from Cauchy sequences, but outside a few files it uses them only
through results such as the field axioms and completeness, so the construction
could be replaced. Finding a characterizing interface strong enough to work with
can itself take new mathematics~\cite{buzzard2024grothendieck}.
Mathlib has refactored fraction fields in this way, replacing a single
construction with a characterizing property~\cite{baanen2022formalization}. A theorem is like a function whose statement is its type and whose proof is its
body, and callers depend only on the statement.

\paragraph{Coupling and cohesion.} \citet{stevens1974structured} judged a
division into modules by its coupling and its cohesion. Coupling is how
strongly modules depend on one another, and it should be low. Cohesion is how
closely the parts of one module belong together, and it should be high. A lemma that states one fact is
fully cohesive, while a long proof whose \code{have} steps each establish a
separate fact bundles several facts that could each be a lemma. In Lean,
coupling shows in a statement's arguments and in how often proofs unfold
definitions. The little theories method of \citet{farmer1992little} applies
a related modular principle to formal mathematics, proving each theorem in the
smallest theory that supports it and reusing it elsewhere through theory
interpretations.

\paragraph{Fan-in, layers, and tails.} Fan-in, the number of modules that use
a given module, is one of the oldest structural measures of software design,
and \citet{henry1981software} built an early metric on a version of it based
on information flow. A part with high fan-in is reused, which is the point of naming it. \citet{parnas1979designing} defined the relation of one
part using another and argued that it should form a hierarchy of levels, each
built on those below. In a Lean library, the uses of a declaration measure its
fan-in, and its dependency height its level. Fan-in is heavy-tailed in
software~\cite{louridas2008power}, with a few parts used everywhere and most
used rarely, so a mean says little. We therefore compare whole distributions
against a reference library, much as software assessment derives quality thresholds from metric
distributions over a benchmark of
systems~\cite{alves2010deriving}.

\paragraph{Refactoring and compression.} The central refactoring of
\citet{fowler1999refactoring} extracts part of a long function into a named function of its own, which shortens the original and removes
duplicated code. In
Lean it turns a \code{have} step into a lemma. \citet{commelin2024abstraction}
make a related point about constructions, which cannot be used outside the
proof that builds them and so, if needed later, ``must be factored out of the
proof into a standalone definition.'' REFACTOR~\cite{zhou2024refactor} learns to make this
extraction automatically, and library learning chooses the abstractions that
most shorten a corpus of
programs~\cite{ellis2021dreamcoder,bowers2023stitch,grand2024lilo}.
\citet{aksenov2026compression} argue that human mathematics is distinguished by
its compressibility through nested definitions, lemmas, and theorems.

\section{What we measure}
\label{sec:measures}

We measure the theorems, definitions, instances, and types that authors wrote in
each library, and leave out metaprogramming, aliases, and declarations generated
by attributes (Section~\ref{sec:design}). Each measurement below captures one
aspect of the structure described in Section~\ref{sec:background}.

\paragraph{Inline claims.} The number of \code{have} and \code{suffices}
steps in a theorem's proof. Each proves a fact that no other declaration can
use. Some such facts belong where they are, but a rate well above that of
human-written code suggests facts that could have been lemmas. We do not count \code{obtain} or
\code{rcases}, which take apart existing facts rather than prove new ones.

\paragraph{Tactic steps.} The number of tactic steps in a declaration's
elaborated proof, including the steps nested inside other steps. It measures
how much work a proof does itself.

\paragraph{Statement and proof length.} Source tokens before and after
\code{:=}, comments excluded. A library of small lemmas has short statements
and short proofs.

\paragraph{Explicit arguments.} The share of theorems with three or more
explicit arguments. Mathlib makes an argument implicit when Lean can infer it
from the others, so that a theorem applies without restating the objects it is
about.

\paragraph{Uses.} The number of other declarations in the same library whose
statement or proof refers to a declaration, which is its fan-in.

\paragraph{Dependency height.} The longest chain of references from a theorem
that stays inside its library.

\paragraph{Instances.} The share of declarations that are type-class
instances. Mathlib organizes mathematics through type classes, so a result
proved for a structure applies to all of its instances.

\paragraph{Private declarations.} The share of declarations marked
\code{private}, which no other file can use.

\section{Libraries and method}
\label{sec:design}

\paragraph{Libraries.} Table~\ref{tab:corpora} lists the six libraries of
the main analysis with the revisions we measured.
We leave out projects aimed at a single theorem, such as the human-written
projects on Fermat's Last Theorem~\cite{buzzard2026flt}, Carleson's
theorem~\cite{vandoorn2026carleson}, and the polynomial Freiman--Ruzsa
conjecture~\cite{tao2026pfr}, together with AI-generated projects of the same
kind. For PhysLib we measure its reviewed libraries, \code{Physlib} and
\code{QuantumInfo}, and leave out \code{PhyslibAlpha}, which accepts
AI-generated contributions under lighter review. FormalFrontier is a collection
of 44 libraries, many of which depend on others in the collection. We treat it
as one library, so its combined size overstates the size of any one of them. Ten of them did not build completely,
missing 32 of the collection's 3,273 modules. One depends on a different
Mathlib revision from the other 43, and we compare all 44 with the revision
the 43 share.

\begin{table}[b]
\centering\small
\caption{The libraries of the main analysis and Mathlib, whose own statistics
we report at the revision that PhysLib and Atlas use. Lines are thousands of
source lines in the library's own files. Declarations are theorems,
definitions, instances, and types that an author wrote.}
\label{tab:corpora}
\begin{tabular}{@{}llrrlll@{}}
\toprule
\thead{Library} & \thead{Written by} & \thead{Lines (k)} & \thead{Declarations} & \thead{Revision} & \thead{Lean} & \thead{Mathlib} \\
\midrule
Mathlib & human & 2,358 & 254,173 & \code{56a00585} & 4.35.0 & -- \\
\rowrule
CSLib & human & 49 & 4,776 & \code{f8acdb37} & 4.36.0-rc2 & \code{887bb849} \\
\rowrule
PhysLib & human & 207 & 13,915 & \code{25465e80} & 4.35.0 & \code{56a00585} \\
\rowrule
FormalFrontier & AI & 713 & 38,365 & 44 libraries & 4.34.0-rc2 & \code{83abb3e7} \\
\rowrule
Tau Ceti & AI & 6,500 & 366,101 & \code{7550358f} & 4.36.0-rc2 & \code{1b1ccf4b} \\
\rowrule
CSLib Frontier & AI & 1,836 & 109,786 & \code{593563c5} & 4.36.0-rc2 & \code{69552756} \\
\rowrule
Atlas & AI & 1,805 & 84,645 & \code{321c3f1a} & 4.35.0 & \code{56a00585} \\
\bottomrule
\end{tabular}
\end{table}

\paragraph{Method.} We build each library under its own toolchain. A Lean
program~\cite{moura2021lean4} loads it with its Mathlib revision and records
each constant's kind, source range, and the constants its statement and proof
refer to. A lexer splits each declaration's source into tokens to measure
lengths and count tactic steps. Constants that Lean generates for a
declaration, such as pieces of its proof and its equation lemmas, count as
part of it. We keep only declarations whose written name is their own, which
drops aliases and declarations generated by attributes such as
\code{to\_additive}, and we leave out metaprogramming. To count steps and
claims in elaborated proofs, a second program elaborates every file again under
the library's own options and records the goal before each tactic step. Every
file elaborates without errors.

\paragraph{Comparison.} We compare each library with the Mathlib revision it
depends on. We also treat Mathlib's 26 top-level areas (``Mathlib areas'') as
separate libraries, which shows how much a statistic varies within
human-written code and how it grows with size. Mathlib's statistics barely change between the
revisions the libraries pin. Reuse is counted only inside a library or area, so CSLib
Frontier's uses of CSLib, for example, do not count. We do not control for subject. We computed several dozen candidate
statistics and report the ones we found most illustrative of the differences.
Appendix~\ref{app:inventory} lists them all.

\paragraph{Reading the figures.} The percentile figures
(Figures~\ref{fig:proofs} and~\ref{fig:stmt_tokens} and
Appendix~\ref{app:distance}) plot a statistic at each percentile of a library's
theorems, and a second panel shows how many percentiles higher each theorem
would rank in Mathlib.

\section{Evidence}
\label{sec:evidence}

\begin{figure}[t]
\centering
\includegraphics[width=\linewidth]{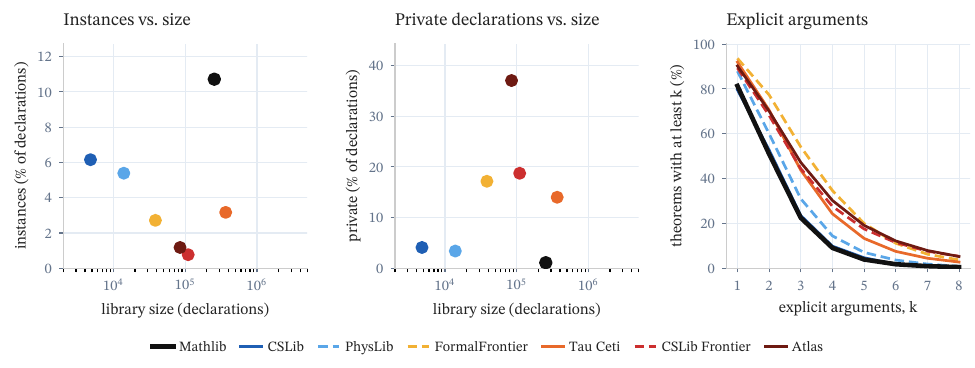}
\caption{Left and middle, the share of declarations that are instances and the
share that are private, against library size. Right, the share of theorems with
at least $k$ explicit arguments.}
\label{fig:headline}
\end{figure}

\begin{table}[b]
\centering\small
\caption{Summary of each library. Text used once is the share of source
tokens in declarations used exactly once in the library. The last column is
the share of theorems with three or more explicit arguments. The Mathlib areas
row gives the range across Mathlib's 26 top-level areas.}
\label{tab:headline}
\begin{tabular}{@{}lrrrrrr@{}}
\toprule
 & \thead{Inline claims} & \thead{Theorems} & \thead{Text} & & & \thead{Explicit} \\
\thead{Library} & \thead{per theorem} & \thead{with a claim} & \thead{used once} & \thead{Instances} & \thead{Private} & \thead{arguments} \\
\midrule
\rawinput generated/headline.tex
\bottomrule
\end{tabular}
\end{table}

\subsection{Local proof structure}

\begin{figure}[t]
\centering
\includegraphics[width=\linewidth]{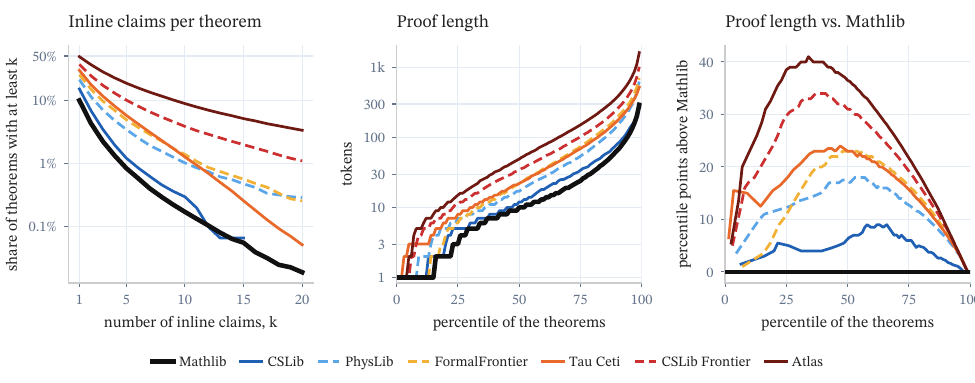}
\caption{Left, the share of theorems with at least $k$ inline claims. Middle,
proof length in tokens at each percentile of theorems. Right, how many
percentiles higher each proof would rank in Mathlib, so a library on the zero
line matches Mathlib.}
\label{fig:proofs}
\end{figure}

Proofs in the AI-generated libraries contain 0.81 to 3.22 inline claims per
theorem, against 0.21 to 0.64 in the human-written ones, and their median proof
is 22 to 49 tokens long against 10 to 17 (Figure~\ref{fig:proofs}). Each inline claim is an intermediate fact proved inside the proof with
\code{have} or \code{suffices}. No other declaration can use it, so reasoning that
could have been a named lemma stays inside the one proof that needed it.

A quarter to a half of the theorems in each AI-generated library state at
least one inline claim, against 10 to 21\% in the human-written ones. In Atlas nearly one theorem in five has five or more, against fewer than one in a
hundred in Mathlib. The difference runs through the whole distribution and is
not confined to a few long proofs.

In every library the theorems that other declarations use carry more inline
claims than those that nothing uses, 0.28 against 0.09 in Mathlib and 3.47
against 2.17 in Atlas, and the AI-generated libraries are above Mathlib in both
groups. A library built to reach target theorems has many theorems that
nothing uses, but its inline claims are not concentrated in them, so the
difference does not come from what the libraries are for.

\begin{figure}[b]
\centering
\includegraphics[width=\linewidth]{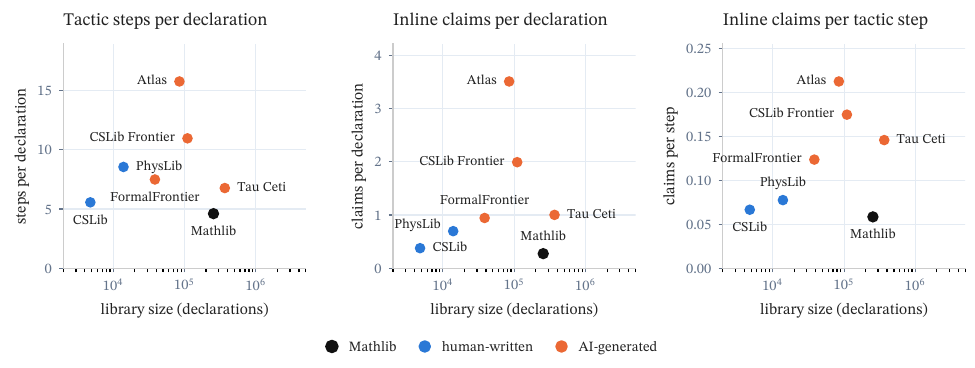}
\caption{Elaborated proofs against library size. Left and middle, tactic steps
and inline claims per declaration with a tactic proof. Right, inline claims per
tactic step. Human-written libraries are blue, AI-generated ones orange, and
Mathlib black.}
\label{fig:goals}
\end{figure}

\begin{figure}[t]
\centering
\includegraphics[width=\linewidth]{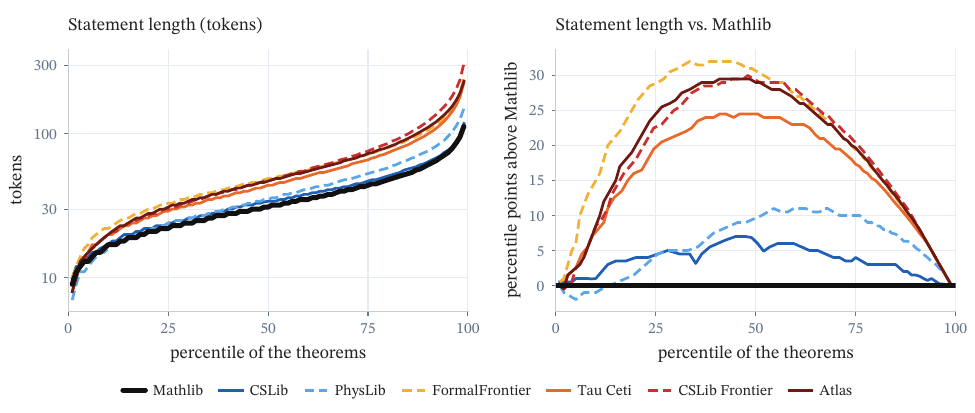}
\caption{Statement length of theorems in source tokens, from the declaration
keyword up to the proof. Left, the length at each percentile of theorems.
Right, how many percentiles higher each statement would rank in Mathlib, so a
library on the zero line matches Mathlib.}
\label{fig:stmt_tokens}
\end{figure}

Per hundred tactic steps, the AI-generated libraries state 12 to 21 new facts
against 6 to 8 in the human-written ones, while the number of steps per
declaration separates the two groups less, 6.8 to 15.8 against 4.6 to 8.5
(Figure~\ref{fig:goals}). The elaborated proofs thus give the same ordering as
the source text, and the difference lies less in how many steps a proof takes
than in how many new facts it states along the way. Statements differ less
than proofs, with a median of 44 to 49 tokens against 31 to 35
(Figure~\ref{fig:stmt_tokens}).

\subsection{Observed reuse}
\label{sec:reuse}

\begin{figure}[b]
\centering
\includegraphics[width=\linewidth]{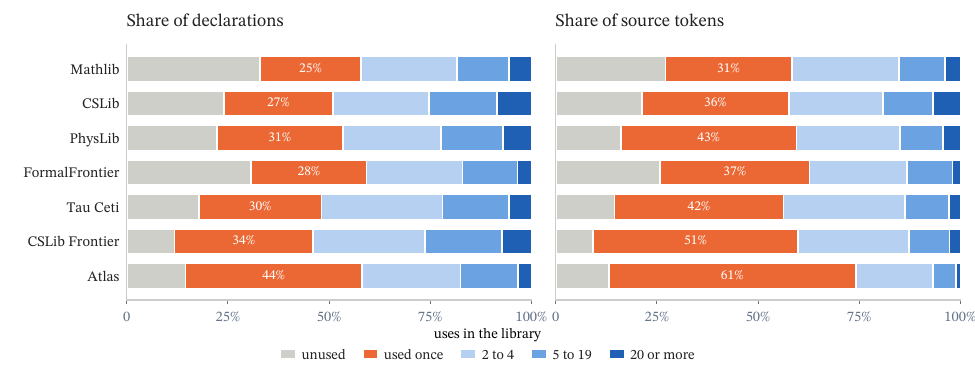}
\caption{How many declarations of the same library use each declaration, as a
share of declarations (left) and of source tokens (right). Labels give the share
used exactly once.}
\label{fig:uses}
\end{figure}

Atlas and CSLib Frontier put half or more of their text into declarations used
exactly once, against a third in Mathlib (Figure~\ref{fig:uses}). Most of these
declarations feed further declarations, forming chains of lemmas each used once.
Proofs are split into steps rather than into parts that other proofs could
share. Mathlib instead has the most declarations that nothing
inside it uses, because a reference library states results for its users.

Among declarations used at least once, the share used at least twice lies within the range of Mathlib's areas in every library except
Atlas. Weighted by length, Atlas and CSLib Frontier fall below every area, so
the declarations they do reuse tend to be short.

The median theorem of Mathlib sits on a chain of 15 references inside its own
area, against 3 to 9 in the six libraries (Figure~\ref{fig:depth_own} in
Appendix~\ref{app:distance}). Height depends on size, on the age of a library,
on where its boundary is drawn, and on how much it takes from upstream, so we
read it as a description of layering and not as a defect.

\subsection{Design choices}

The AI-generated libraries register 0.8 to 3.2\% of their declarations as
instances against 5.4 to 10.7\% in the human-written ones, mark 14 to 37\%
private against at most 4\%, and give about half of their theorems many
explicit arguments against a quarter to a third (Figure~\ref{fig:headline} and
Table~\ref{tab:headline}). These are choices an author makes when declaring
things rather than when proving them, and each affects how easily other code
can use a result. Fewer instances mean that fewer general results apply to
their structures automatically, a private declaration cannot be used from
another file, and explicit arguments must be supplied by hand. On private
declarations and explicit arguments every AI-generated library lies outside the
range of Mathlib's areas. None of the three depends on proof style.

\subsection{Authorship and purpose}
\label{sec:purpose}

On inline claims, proof length, and instances, the two human-written libraries
are nearest to Mathlib and the four AI-generated ones are further. Tau Ceti and
FormalFrontier are the closest of the AI-generated libraries on proof
structure, but they register instances at three fifths of the rate of the
human-written libraries or less.

CSLib has 0.31 inline claims per theorem, a median proof of 12 tokens, 6.2\%
instances, and 4.1\% private declarations, against 1.61, 34 tokens, 0.8\%, and
18.7\% in CSLib Frontier, which covers the same subject and builds on CSLib.
The two differ in size and in aim as well as in authorship.

Mathlib's \code{Archive} and \code{Counterexamples}, one-off proofs that the
Mathlib community writes and reviews outside the main library, have 0.89 inline
claims per theorem against Mathlib's 0.21, and 31\% of their theorems state at
least one claim against 10\%. On proof structure they look like the
AI-generated libraries, which suggests that purpose, and not only authorship,
shapes it. On instances (6.4\%), private declarations (3.3\%), and theorems
with three or more explicit arguments (19\%), however, they are like Mathlib,
so purpose does not explain these three differences.

With two human-written and four AI-generated libraries, these observations describe the libraries studied and do not estimate an effect
of authorship. Studies of AI coding tools in software report increased complexity and less
refactoring~\cite{he2026speed,harding2024coding}, in the same direction as our
observations.

\section{Limitations}
\label{sec:limitations}

\paragraph{Alternative explanations.} Subject matters, since Mathlib's own areas
differ by a factor of seventeen on inline claims. Purpose may matter as much as
authorship for proof structure, as the Archive suggests
(Section~\ref{sec:purpose}). Mathlib has
been refactored over many years, while the six libraries are younger and less
maintained, so age may explain part of the gap. File and library
boundaries are conventions, and FormalFrontier, a collection of 44 libraries,
has no single boundary. Finally, every \code{have} or
\code{suffices} counts as an inline claim, so a one-line \code{have} that feeds
automation counts as much as a substantial one.

\paragraph{Validation.} We have not tested whether the statistics respond to
refactoring as they should. Material that a project moves into Mathlib, as
large projects routinely do~\cite{vandoorn2023sphere}, should move towards
Mathlib's values, and inlining lemmas in a Mathlib area should move each
statistic away from them.

\section{Related work}

\paragraph{Statistics of proof corpora.} \citet{blanchette2015mining} mined
the Archive of Formal Proofs for size and dependency statistics,
\citet{li2026network} study the dependency graph of Mathlib, and
\citet{aspinall2016metrics} adapt software design metrics to proof libraries.
Few studies compare human-written with AI-generated mathematics.
\citet{li2026ai} find that language models reproving the propositions of
Euclid's \emph{Elements} cite foundational elements, such as definitions and
postulates, far more often than human proofs do, an informal analogue of
proving facts inside the proof. \citet{aryal2026prooftheory} compares
human and AI-generated proofs by measures inside individual proof terms, while
our measurements concern reuse across a library. \citet{berlotattwell2024library}
find that the lemma and function libraries built by two language-model
library-learning systems are rarely reused.

\paragraph{Judging and improving proofs.}
ProofJudge~\cite{caldwell2026proofjudge} scores proofs with a language model,
and the Lean quality
indicators~\cite{degenne2026indicators,morrison2026simple} use one to flag
unnecessary hypotheses and duplicated material in Mathlib, Tau Ceti,
FormalFrontier, and other libraries. The indicators also
include static counts over whole libraries, such as how often definitions are
unfolded outside their own files, which is close to our view of coupling.
Proof-Refactor~\cite{fu2026proofrefactor} extracts lemmas from proofs, and
ProofOptimizer~\cite{gu2026proofoptimizer} shortens generated proofs. The
judged methods work one declaration or file at a time. Distributional
measurements need no judge and are cheap to reproduce, but they describe a
library and not a declaration, and their reference is itself the product of
human review.

\section{Conclusion}

AI-generated Lean libraries differ from human-written ones in how proofs are
structured and how declarations are reused and declared, and the differences
show in simple distributions computed from the source and the dependency graph.
Their proofs state more intermediate facts and run longer, more of their text
sits in declarations used once, and they register fewer instances and mark more
declarations private. One-off proofs written by Mathlib's community share the
proof structure of the AI-generated libraries but not their design choices, so
purpose may explain part of the first difference but not the second. Whether
these statistics track what reviewers and maintainers mean by quality remains
open.

\FloatBarrier
{\small\raggedright
\bibliographystyle{plainnat}
\bibliography{references}
}

\appendix

\section{Candidate statistics}
\label{app:inventory}

Table~\ref{tab:inventory} lists the candidate library-level statistics from
which the main-text measurements were chosen. ``Trend'' is the correlation,
across Mathlib areas, between the statistic and the logarithm of the number of
declarations.

\paragraph{Candidates we did not report.} On fixed-size samples,
source compressibility mainly measures how much of a library a sample covers.
Token-level distance from Mathlib (cross-entropy under a trigram model, and
compression given Mathlib as context) mainly measures vocabulary, since CSLib
and PhysLib are as far from Mathlib as any library. Statistics of elaborated
terms, including a variant of the unwrapped length of \citet{aksenov2026compression}, vary
as much among Mathlib's areas as among the libraries. The share of
declarations used only in their own file depends on file conventions. An
aggregate distance over many statistics loses the direction of each
difference, so we report per-family distances only in
Appendix~\ref{app:distance}.

\paragraph{Measurements that need the elaborated proof.} Several sharper
statistics require more of the elaborated proof than the counts of
Figure~\ref{fig:goals}. One is the
number of inline claims whose statement does not depend on the local context,
so that each could stand alone as a lemma. Others are how often proofs unfold a
definition where an interface lemma would serve, and how many subterms are
duplicated across proofs up to renaming.

\begin{table}[p]
\centering
\caption{Candidate statistics. FF is FormalFrontier and CSLib Fr.\ is CSLib Frontier.}
\label{tab:inventory}
\footnotesize\setlength{\tabcolsep}{3pt}
\makebox[\linewidth][c]{%
\begin{tabular}{@{}>{\raggedright\arraybackslash\hangindent=2em\hangafter=1}p{2.15in}rrrrrrrrr@{}}
\toprule
\thead{Statistic} & \thead{Mathlib} & \shortstack[r]{\textbf{Mathlib}\\\textbf{areas}} & \thead{CSLib} & \thead{PhysLib} & \thead{FF} & \shortstack[r]{\textbf{Tau}\\\textbf{Ceti}} & \shortstack[r]{\textbf{CSLib}\\\textbf{Fr.}} & \thead{Atlas} & \thead{Trend} \\
\midrule
\rawinput generated/inventory.tex
\bottomrule
\end{tabular}}
\end{table}

\section{Further distributions and distances}
\label{app:distance}

Figures~\ref{fig:type_dag} to~\ref{fig:taste_log2} show the remaining
per-theorem statistics that vary smoothly, for the six libraries, grouped by the
families of Table~\ref{tab:distance} and each with the panels of
Figure~\ref{fig:stmt_tokens}. Table~\ref{tab:distance} gives, for
each family of per-declaration statistics, the median over the family of a
distance ratio. The library's values and Mathlib's are expressed as percentiles of Mathlib, compared
quantile by quantile, and divided by the median of the same distance between
each Mathlib area and the rest of Mathlib. A ratio of 1 is the distance of a
typical Mathlib area. The ratio has no direction and depends on the partition
into areas.

\begin{table}[H]
\centering\small
\caption{Distance ratio by family of per-declaration statistics.}
\label{tab:distance}
\begin{tabular}{@{}lrrrr@{}}
\toprule
\thead{Library} & \thead{Signature} & \thead{Proof} & \thead{Reuse} & \thead{Compression} \\
\midrule
\rawinput generated/tables.tex
\bottomrule
\end{tabular}
\end{table}

\rawinput generated/appendix-figures.tex

\end{document}